%% file: 000main.tex
\documentclass{article}

\PassOptionsToPackage{numbers}{natbib}

    \usepackage[preprint]{neurips_2022}

\usepackage[utf8]{inputenc} 
\usepackage[T1]{fontenc}    
\usepackage{hyperref}       
\usepackage{url}            
\usepackage{booktabs}       
\usepackage{amsfonts}       
\usepackage{nicefrac}       
\usepackage{microtype}      

\usepackage[table,xcdraw,dvipsnames]{xcolor}
\usepackage{graphicx} 
\newcommand{\sufx}[1]{$_{\text{#1}}$}
\newcommand{\red}[1]{\textcolor{BrickRed}{#1}}

\usepackage{amsmath,amsfonts,amssymb}
\usepackage{cleveref}
\crefformat{section}{\S#2#1#3} 
\crefformat{subsection}{\S#2#1#3}
\crefformat{subsubsection}{\S#2#1#3}

\input{math_commands}

\title{Latent Topology Induction for Understanding Contextualized Representations}

\author{%
  Yao Fu\quad\quad\quad\quad Mirella Lapata\\
  Institute for Language, Cognition and Computation\\
  University of Edinburgh\\
  yao.fu@ed.ac.uk \quad\quad\quad\quad mlap@inf.ed.ac.uk \\
}

\begin{document}

\maketitle

\begin{abstract}
  In this work, we study the
  representation space of contextualized embeddings 
  and gain insight into the hidden topology of large language models.
  We show there exists a network of latent states that summarize
   linguistic properties of contextualized representations. 
  Instead of seeking alignments to existing well-defined annotations,
  we infer this latent network in a fully unsupervised way
  using a structured variational autoencoder.
  The induced states not only serve as anchors that mark the topology (neighbors and connectivity) of the representation manifold
  but also
  reveals the internal mechanism of encoding sentences.
    With the induced network, we: 
  (1). decompose the representation space into a spectrum of latent states which encode fine-grained word meanings with lexical, morphological, syntactic and semantic information; 
  (2). show state-state transitions encode rich phrase constructions and serve as the backbones of the latent space. 
  Putting the two together, we show that sentences are represented as a traversal over the latent network where state-state transition chains  encode syntactic templates and state-word emissions fill in the content. 
  We demonstrate these insights with extensive experiments and visualizations. 
\end{abstract}

\section{Introduction}
\label{sec:intro}
\input{010intro}

\input{fig_state_word}

\section{Background}
\input{020background}

\section{Method}
\label{sec:method}
\input{030method}

\input{tab_overall}

\input{tab_not_aligned}
\input{tab_lex_sem_def}
\section{Experimental Setting}
\label{sec:exp_setting}
\input{040exp_setting}

\section{State-Word Topology}
\label{sec:state-word}
\input{050state_word}

\section{State-State Topology}
\label{sec:state-state}
\input{060state_state}

\section{Conclusions}
\label{sec:conclusion}
\input{070conclusion}

\bibliographystyle{plainnat}
\bibliography{latent_network}



\appendix
\input{app_main.tex}



\end{document}

%% file: math_commands.tex
\usepackage{amsmath,amsfonts,bm}

\def\eqref#1{equation~\ref{#1}}

\def\1{\bm{1}}

\def\vh{{\bm{h}}}

\def\vr{{\bm{r}}}
\def\vs{{\bm{s}}}

\def\vx{{\bm{x}}}

\def\vz{{\bm{z}}}

\DeclareMathAlphabet{\mathsfit}{\encodingdefault}{\sfdefault}{m}{sl}
\SetMathAlphabet{\mathsfit}{bold}{\encodingdefault}{\sfdefault}{bx}{n}



%% file: 010intro.tex
 
Recently, there has been considerable interest in analyzing pretrained language models (PLMs)~\citep{rogers-etal-2020-primer, hewitt-manning-2019-structural, hewitt-liang-2019-designing, chen2021probing,chi-etal-2020-finding,  liu-etal-2019-linguistic} due to their huge success.
This work aims to investigate the intrinsic geometric and topological properties of contextualized representations.
We study how words within sentences, as sequences of vectors, are positioned within the representation space. 
We hypothesize that there exists a spectrum of
latent anchor embeddings (or landmarks) that describe the manifold
topology.
As a quick first impression, 
Figure~\ref{fig:state_word} shows the latent states that 
we will discover in the following sections. 
Since such structure cannot be straightforwardly observed, we infer
the topology as latent variables.

Our approach does not follow the majority of previous  literature
on probing which usually defines a probe as a classifier
\textit{supervised} by existing annotations like part-of-speech~\citep{mamou2020emergence}, dependency trees~\citep{hewitt-manning-2019-structural}, CCG supertags~\citep{liu-etal-2019-linguistic} and others~\citep{tenney-etal-2019-bert, chi-etal-2020-finding}.
We do not assume interpretations of the latent topology will
be strictly aligned with well-defined linguistic annotations, as
they simply do not have to.
The dilemma of supervised probing, 
as pointed out by much previous work~\citep{hewitt-liang-2019-designing, hall-maudslay-etal-2020-tale, chen2021probing}, is that
 it is hard to differentiate whether the discovered linguistic properties are intrinsically within the model, or imposed by the supervision signal. 
In this work, to maximally reduce external bias, we infer the latent states in a fully unsupervised way.
As long as the inferred states are intrinsically meaningful (see Fig.~\ref{fig:state_word} for example states), it does 
 not matter whether they align with well-defined annotations or not. 

We use a structured variational autoencoder (VAE)~\citep{kingma2013autoencoding} to infer the latent topology, as VAEs are common and intuitive models for learning latent variables.
We focus on the manifold where contextualized embeddings lay in
(e.g., the last layer outputs of a fixed, not fine-tuned,
BERT~\citep{devlin-etal-2019-bert}). 
We hypothesize there exists a wide spectrum of \textit{static} latent states within this manifold and
 assume two basic generative properties of the states: 
(1). a state should summarize the meaning of its corresponding words and contexts;
(2). transitions between different states should encode sentence structure. 
We model these two properties as emission and transition potentials of a CRF~\citep{sutton2012introduction} inference network.
Since a VAE is a generative model trained by reconstructing sentences, essentially we learn states that are informative enough to generate words and sentences. 

The first part of our experiments show how states encode linguistic properties of words (\cref{sec:state-word}). 
We show that states summarize rich linguistic phenomena ranging from lexicon, morphology, syntax to semantics.  
We highlight two intriguing effects of contextualization:
(1). before contextualization, function words (e.g., \textit{to, for, and, or} .etc) are concentrated around few top states;
contextualization spread function words over the whole space.
(2). before contextualization, certain tokens do not form meaningful states; contextualization makes them ``receive'' meaning from their neighbors.

The second part of our experiments show that transitions 
between states form the topological backbone of the representation
space and provide further grounding for sentence construction (\cref{sec:state-state}). 
We differentiate two types of states within the space: states 
encoding \textit{function words} and states encoding \textit{content words}.
We show that function states serve as ``hubs'' in state transitions and
attract content words of similar meanings to be close.
State transitions then encode rich types of phrase constructions. 
Finally, putting everything together, our most important 
discovery is \textit{a step-by-step
latent mechanism} about how sentences are represented as a  traversal
over the discovered topology (\cref{sec:state_state:sentence_traversal}).

%% file: fig_state_word.tex
\begin{figure*}[t!]
\small
  \centering
  \includegraphics[width=\linewidth]{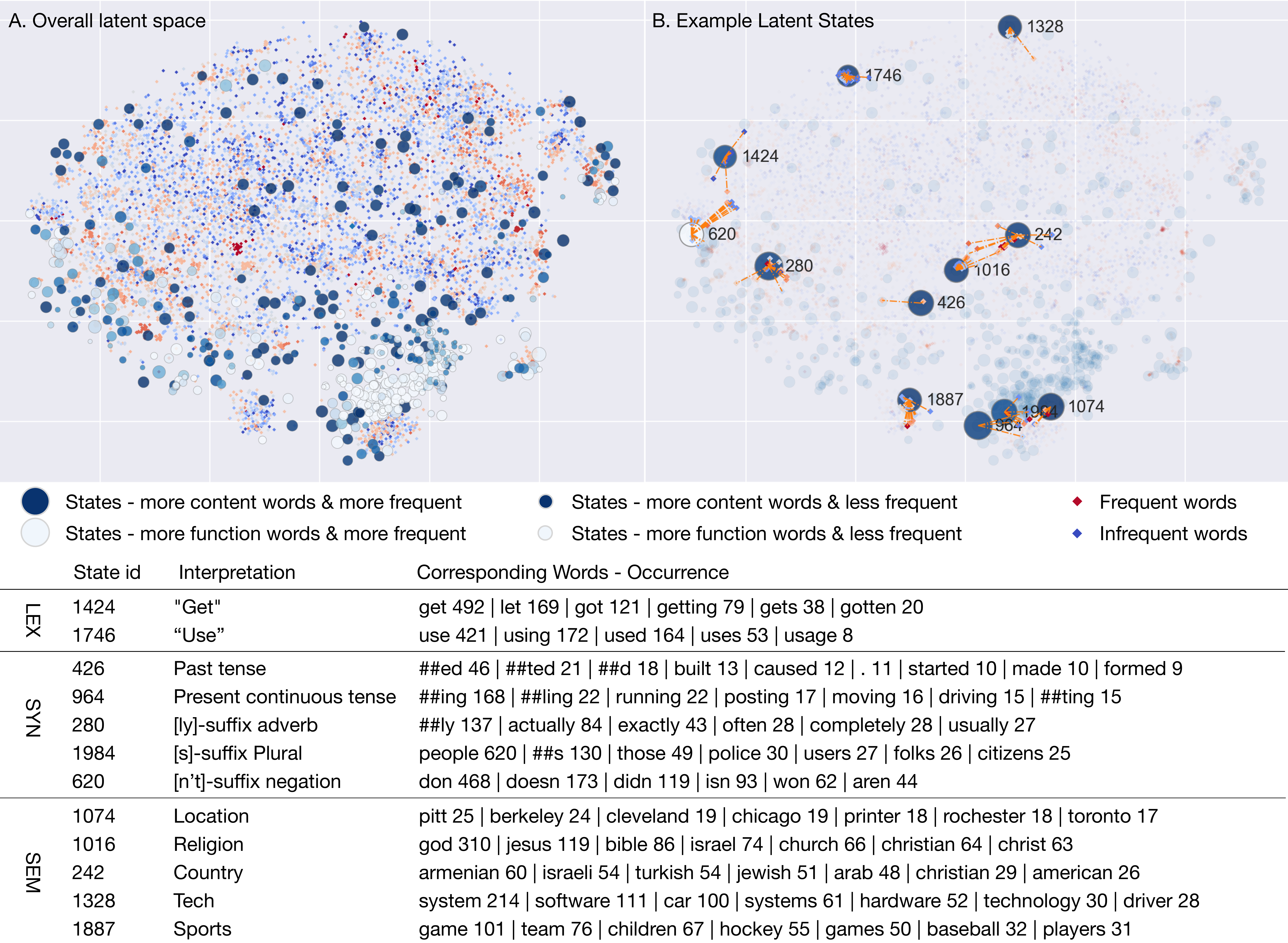}
  \caption{We discover a spectrum of  latent states with lexical, morphological, syntactic and semantic interpretations.
  The states also summarize the topological structure of the representation space of language models. 
  See \cref{sec:state-word} and \cref{sec:state-state} for how these states are discovered. 
  }
  \label{fig:state_word}
\end{figure*}

%% file: 020background.tex
\textbf{Supervised Probing}\quad 
Collectively known as
``Bertology''~\citep{hewitt-manning-2019-structural, 
rogers-etal-2020-primer}, the goal of probing is to 
discover what is \textit{intrinsically} encoded within large language models.
The mainstream approach is to construct a supervised weak classifier (a.k.a. a probe) and fine-tune it with classical linguistic annotations, like part-of-speech tagging~\citep{liu-etal-2019-linguistic, hewitt-etal-2021-conditional}, edge detection~\citep{tenney2018what, tenney-etal-2019-bert}, parsing~\citep{hewitt-manning-2019-structural, hewitt-liang-2019-designing, NEURIPS2019_159c1ffe, chi-etal-2020-finding} and sentiment classification~\citep{chen2021probing, wu-etal-2020-perturbed}.
The problem here is that it is difficult to tell if 
the discovered properties are intrinsic to the embedding or imposed by the supervision signal~\citep{hewitt-liang-2019-designing, chen2021probing, hall-maudslay-etal-2020-tale}.
Since our method is fully unsupervised, our results are more intrinsic w.r.t. the model. 

\textbf{Unsupervised Methods}\quad
To bypass issues with supervised probing,
some unsupervised work proposes to extract syntactic~\citep{Kim2020Are}, geometric~\citep{cai2021isotropy}, cluster-based~\citep{dalvi2022discovering}, and other information
from PLMs~\citep{wu-etal-2020-perturbed, michael-etal-2020-asking}.
Our focus here is the network topology within the representation space, which is not yet thoroughly studied. 
Amongst the large volume of Bertology research, the closest unsupervised work to ours are: \citet{dalvi2022discovering} who use clustering to discover latent concepts within BERT, and we will later use their results as a comparison to our discoveries; 
\citet{michael-etal-2020-asking} who discovers latent ontology in an unsupervised way;
\citet{cai2021isotropy} who study the geometric properties of BERT with a focus on isotropy. 
There is also supervised method for extracting static embeddings from contextualized embeddings~\citep{gupta-jaggi-2021-obtaining}.
These work more or less involve cluster structures within BERT. 
Our major difference is that we take an important further step from
word clusters to state-state transitions and show how traversal
over states leads to sentence constructions. 
In the latent variable literature, our inference model uses
a classical CRF-VAE formulation~\citep{Fu2020GumbelCRF, mensch2018differentiable}.
Existing work uses this formation for other tasks like structured 
prediction~\citep{ammar2014crfautoencoder} or sentence generation~\citep{li-rush-2020-posterior} while we discover latent structures within PLMs.

%% file: 030method.tex

\textbf{Latent States within Representation Space}\quad 
Given a sentence $\vx = [x_1, ..., x_T]$, 
we denote its contextualized representations as $[\vr_1 , ..., \vr_T] = \text{PLM}(\vx)$ where PLM$(\cdot)$ denotes a pretrained encoder (here we use BERT and our method is applicable to any PLM).
Representations $\vr$ for all sentences lie in one manifold $\mathcal{M}$, 
namely the representation space of the language model. 
We hypothesize there exists a set of $N$ static latent states 
$\vs_1,...,\vs_N$ 
that function as anchors and outline the space topology (recall Fig.~\ref{fig:state_word}).
We emphasize that all parameters of the PLM are fixed 
(i.e., no fine-tuning), 
so all learned states are intrinsically within $\mathcal{M}$. 
We focus on two topological relations: 
(1). state-word relations, which represent how word embeddings may be summarized by their states and how states can be explained by their corresponding words; 
(2). state-state relations, 
which capture how states interact with each other and how their transitions denote meaningful word combinations.
Taken together, these two relations form a latent network within $\mathcal{M}$ (Fig.~\ref{fig:state_word} and later Fig.~\ref{fig:transition}).

\textbf{Modeling}\quad 
For state-word relations, 
we associate each word embedding $\vr_t$ 
with a latent state indexed by $z_t \in \{1, ..., N\}$.
We use an emission potential $\phi(x_t, z_t)$ to model how $z_t$ is likely to summarize $x_t$. 
The corresponding embedding of $z_t$ is then $\vs_{z_t}$. 
For state-state relations, we assume a transition matrix $\Phi(z_{t-1}, z_t)$ modeling the affinity about how state $z_{t-1}$ are likely to transit to state $z_t$. 
Together $\phi(x_t, z_t)$ and $\Phi(z_{t-1}, z_t)$ form the potentials of a linear-chain CRF:
\begin{align}
    \log \phi(x_t, z_t) = \vr_t^\intercal \vs_{z_t} \quad\quad\quad\quad \log \Phi(z_{t-1}, z_t) = \vs_{z_{t-1}}^\intercal \vs_{z_{t}}
\end{align}
where the vector dot product follows the common practice of fine-tuning contextualized representations.
The probability of a state sequence given a sentence is:
\begin{align}
    q_\psi(\vz|\vx) = \prod_{t=1}^T\Phi(z_{t-1}, z_t)\phi(x_t, z_t) / Z
\end{align}
where $Z$ is the partition function. 
Note that only embeddings of states $\psi = [\vs_1, ..., \vs_N]$ are learnable parameters of the inference model $q_\psi$. 
To infer $\vs$,
we use a common CRF-VAE architecture shared by previous
work~\citep{ammar2014crfautoencoder, li-rush-2020-posterior, fu2022scaling} and add a generative model on top of the encoder:
\begin{align}
    &p_\theta(\vx, \vz) = \prod_t p(x_t | z_{1:t}) \cdot p(z_t | z_{1:t-1}) \quad \quad 
    \vh_t = \text{Dec}(\vs_{z_{t-1}}, \vh_{t-1}) \\
    &p(z_t | z_{1:t-1}) = \text{softmax}(\text{FF}([\vs_{z_{t}} ; \vh_t])) \quad \quad p(x_t | z_{1:t}) = \text{softmax}(\text{FF}(\vh_t))
\end{align}
where $\theta$ denotes the decoder parameters, Dec$(\cdot)$ denotes
the decoder (we use an LSTM), $\vh_t$ denotes decoder states, and
FF$(\cdot)$ denotes a feed-forward network.  
We optimize the $\beta$-ELBO objective:
\begin{align}
    \mathcal{L}_\text{ELBO} = \mathbb{E}_{q_\psi(z | x)}[\log p_\theta(x, z)] - \beta \mathcal{H}(q_\psi(z | x)) \label{eq:elbo}
\end{align}
%
We further note that the decoder's goal is for help inducing the latent states, rather than being a powerful sentence generator. 
After training, we simply drop the decoder and only look at the inferred states. 
Maximizing $p(x_t | z_{1:t})$ encourages $z_{1:t}$ (thus their embeddings $s_{z}$) to reconstruct the sentence
and $p(z_t | z_{1:t-1})$ encourages previous $z_{1:t-1}$ to be predictive to the next $z_t$ (so we can learn transitions). 
Essentially, this formulation trys to find ``generative'' states $\vs$ within the representation space $\mathcal{M}$ that are able to predict sentences and their next states.

%% file: tab_overall.tex
\begin{table*}[t!]
  \small
  \caption{\label{tab:overall}Decomposing representation space by aligning to existing linguistic annotation. \#s = number of states, \%c = percentage of covered word occurrence. \colorbox[HTML]{FFC897}{POS} and \colorbox[HTML]{FFC897}{ENT} can be inferred from the word directly while
   \colorbox[HTML]{DAE8FC}{DEP}, \colorbox[HTML]{DAE8FC}{CCG} and  \colorbox[HTML]{DAE8FC}{BCN} require more context.  
  Generally, there are more aligned states inferred from \textsc{BertLast}, but they cover fewer word occurrences than \textsc{BertZero}. 
  See Table~\ref{tab:not_aligned} for interpretations of non-aligned states.
  }
  \begin{center}
  \small
  \begin{tabular}{@{}lcccccccccccc@{}} 
  \toprule
  &  \multicolumn{2}{c}{\colorbox[HTML]{FFC897}{POS}}  & \multicolumn{2}{c}{\colorbox[HTML]{FFC897}{ENT}} 
  & \multicolumn{2}{c}{\colorbox[HTML]{DAE8FC}{DEP}} & \multicolumn{2}{c}{\colorbox[HTML]{DAE8FC}{CCG}} & \multicolumn{2}{c}{\colorbox[HTML]{DAE8FC}{BCN}} &  \multicolumn{2}{c}{Not Aligned}  \\ 
  & \#s & \%c & \#s & \%c & \#s & \%c & \#s & \%c  & \#s & \%c & \#s & \%c \\
  \midrule
  \textsc{Positional} & 3 & 0.44 & 0 & 0 & 3 & 0.44 & 1& 0.01 &1 & 0.01 & 1997 & 99.55 \\
  \textsc{RandEmb} & 392 & 53.02 & 21 & 1.06 & 366 & 52.36  & 241 & 45.20 & 334 & 49.16 & 1159 & 43.45 \\ 
  \textsc{BertZero} & 673 & 65.31 & 37 & 2.11 & 468 & 53.84 & 450 & 51.84 & 275 & 48.89 & 1253 & 27.69 \\ 
  \textsc{BertLast} & 804 & 53.23 & 51 & 0.82 & 740 & 42.62 & 583 & 44.03 & 628 & 44.52 & 1069 & 40.52 \\ 
  \bottomrule
  \end{tabular}
  \end{center}
\end{table*}

%% file: tab_not_aligned.tex
\begin{table*}[t!]
  \small
  \caption{\label{tab:not_aligned} Human evaluation (averaged over 3 annotators) of states not aligned with existing annotations.
  We note that most of them are still meaningful.
  \textsc{BertZero} covers more lexical information while \textsc{BertLast} covers more semantic information. See Table~\ref{tab:lex_sem_def} for example lexical (LEX), morphosyntax (SYN) and semantic (SEM) states. 
  }
  \begin{center}
  \begin{tabular}{@{}lcccccccc@{}} 
  \toprule
   &  \multicolumn{2}{c}{\colorbox[HTML]{FFC897}{LEX}}  & \multicolumn{2}{c}{\colorbox[HTML]{DAE8FC}{SYN}} 
  & \multicolumn{2}{c}{\colorbox[HTML]{DAE8FC}{SEM}} & \multicolumn{2}{c}{Not Interpretable}  \\ 
  & \#s & \%c & \#s & \%c  & \#s & \%c & \#s & \%c \\
  \midrule
  \textsc{BertZero} &  83 & 30.37 & 54 & 20.96 & 135 & 37.31 & 39  & 11.37  \\ 
  \textsc{BertLast} &  26 & 15.93 & 40 & 26.25 & 56  & 43.50  & 22  & 14.33  \\ 
  \bottomrule
  \end{tabular}
  \end{center}
\end{table*}

%% file: tab_lex_sem_def.tex
\begin{table*}[t!]
  \small
  \caption{\label{tab:lex_sem_def} Examples of states that are not aligned to existing annotations. N.I. means not interpretable. Subscripts denotes occurrence (e.g, ``win\sufx{69}'' means ``win'' occurs 69 times within the latent state).
  }
  \begin{center}
  \small
  \begin{tabular}{@{}p{0.45cm}ll@{}} 
  \toprule
   & Example states from \textsc{BertZero} & Explanation\\ 
   \midrule
  LEX & win\sufx{69} | won\sufx{26} | wins\sufx{19} | winning\sufx{18} | winbench\sufx{6} | winword\sufx{4} & Most words stem from ``win'' \\
  & utexas\sufx{11} | ah\sufx{10} | umich\sufx{9} | umd\sufx{9} | uh\sufx{9} | udel\sufx{9} | um\sufx{8} | umn\sufx{7} & Most words start with ``u'' \\
  SYN & against\sufx{141} | near\sufx{35} | among\sufx{34} | towards\sufx{27} | toward\sufx{24} | unto\sufx{11} & All are prepositions \\
  & me\sufx{398} | them\sufx{54} | him\sufx{50} | person\sufx{28} & Most are pronouns\\ 
  SEM  & information\sufx{162} | say\sufx{147} | said\sufx{91} | saying\sufx{55} | says\sufx{33} | statement\sufx{31}  & Abount communication \\
  & buy\sufx{78} | sell\sufx{66} | bought\sufx{43} | cheap\sufx{37} | sold\sufx{36} | market\sufx{30} | expensive\sufx{21} & About business \\ 
  N.I.  & course\sufx{37} | however\sufx{26} | way\sufx{25} | know\sufx{14} | said\sufx{12} | yes\sufx{12} & Intuitively not very related\\ 
  & sort\sufx{56} | definition\sufx{19} | kinds\sufx{16} | guilty\sufx{11} | types\sufx{11} | symptoms\sufx{9} & Intuitively not very related\\ 
  \bottomrule
  \end{tabular}
  \end{center}
\end{table*}

%% file: 040exp_setting.tex
\textbf{Dataset, Model Parameters, and Learning}\quad
We perform experiments on the 20News dataset~\citep{kusnerb15documentdist}, 
a common dataset for latent variable modeling 
(initially for topic modeling) as our testbed. 
We primarily focus on the last layer output of a BERT-base-uncased\footnote{\url{https://huggingface.co/bert-base-uncased}} model (\textsc{BertLast}), yet our method is applicable to any larger size, GPT-styled, or encoder-decoder-styled models. 
In terms of model parameters, the dimension of the states is the
same as contextualized embeddings, which is 768.
We use a light parameterization of the decoder and set its
hidden state dimension to 200.
Again, the purpose of the decoder is to help induce the states, rather than being a powerful sentence generator. 
We set the number of latent states $N$ to 2000. 
Recall that the vocabulary size of uncased BERT is 30522, 
which means that if uniform each state approximately
corresponds to 15 words, serving as a type of ``meta'' word. 
We further note that setting $N=2000$ is somehow a sweet point
according to our initial experiments: larger $N$ (say 10K) is too fine-grained and under-clusters (words of
similar linguistic roles are divided into different states)
while smaller $N$ (say 100) over-clusters (words of different roles are
gathered into the same cluster).  
Gradient-based learning of CRFs inference model is challenging due to the intermediate discrete structures. 
So we use approximate gradient and entropy from~\citet{fu2022scaling} 
which enables memory-efficient differentiable training of our model. 
During training, we tune the $\beta$ parameter in Eq.~\ref{eq:elbo}
to prevent posterior collapse, which is a standard training technique for VAEs. 
All experiments are performed on Nvidia 2080Ti GPUs.

\textbf{Baseline Embeddings}\quad
We compare against: 
(1). \textsc{Positional} states induced from positional embeddings. As there is no content information, we expect the induced  structures to be very poor. 
(2). \textsc{RandEmb}, fixed random state embeddings sampled from a 
Gaussian distribution sharing the same mean and variance with \textsc{BertLast} embeddings. 
(3). \textsc{BertZero}, static word embeddings from the zeroth layer of BERT. 
We are particularly interested in the comparison between
\textsc{BertZero} and \textsc{BertLast}, as the differences can
shed light on what happens after contextualization.

%% file: 050state_word.tex
Our study on state-word relations has two steps. 
In~\cref{sec:state_word:decompozation}, we decompose the
interpretation of the inferred states. 
They may align, or not align, with existing annotations. 
We try to align the states first, then use human evaluation to
show even states are not aligned, they still encode lexical/ syntactic/ semantic information. 
In~\cref{sec:state_word:contextualization}, we focus on the effect of contextualization, and show how tokens ``not understood'' in \textsc{BertZero} become ``understood'' in \textsc{BertLast} after contextualization.
We discuss the topology of state-state transitions in \cref{sec:state-state}.

\subsection{Decomposing the Meaning of Inferred States}
\label{sec:state_word:decompozation}
We align the inferred states with:
(1). POS, part of speech tags;
(2). ENT, named entities;
(3). DEP, labels of dependency edges linking a word to its dependency head;
(4). CCG, CCG supertags which contain rich syntax tree information and are usually referred as ``almost parsing''\citep{liu-etal-2019-linguistic};
(5). BCN, BERT Concept Net from~\citet{dalvi2022discovering}, a hierarchy of concepts induced from BERT, mostly about semantics and similar to a topic model. 
We obtain the POS, ENT, DEP annotations on our 20News dataset using a pretrained parser.\footnote{\url{https://spacy.io/}}
We obtain the CCG\footnote{\url{https://groups.inf.ed.ac.uk/ccg/ccgbank.html}} and BCN\footnote{\url{https://neurox.qcri.org/projects/bert-concept-net.html}} annotations from their own websites. 
After training, we use Viterbi decoding to decode states, 
and each state may correspond to multiple copies of 
the same word from different contexts. 
We say a latent state aligns with a predefined tag if 90\% of 
the word occurrence corresponds to this state also corresponds to a tag.
For example, suppose state-0 occurs 100 times in the full validation set, 
90 times of which correspond to either ``happy'' or ``sad'' and 10
times correspond to other random words.
In this case, ``happy'' and ``sad'' takes the dominant
portion of state-0 (90 out of 100), and both are adjectives, 
so we say
state-0 aligns with the POS tag adjective.
We set the threshold to 90\% because it is intuitively high enough.
We select the model that has the largest number of aligned tags to 
the union of the five types of annotations during validation.
Note that this model selection strategy introduces a slight bias, 
yet such bias is much weaker than tuning five separately supervised probes.
We report all our results on the validation dataset.

Table~\ref{tab:overall} shows the alignment results.
First, the results from \textsc{RandEmb} should be viewed as the
intrinsic bias from the modeling power of the LSTM decoder, since
it reconstructs the sentence with even random state embeddings.
This should not be surprising because previous work also reports
neural network's ability to fit random data~\citep{NEURIPS2020_e4191d61, zhang2021understanding}. 
Then we observe more aligned
states inferred from \textsc{BertLast}, but they cover fewer word occurrences than \textsc{BertZero}.
For states not aligned with existing annotations, we ask human
annotators (three graduate students with 100+ TOEFL test scores) 
to annotate if they are:
(1). LEX: words that are textually similar. 
(2). SYN: words share similar morphological-syntactic rules. 
(3). SEM: words with related meaning. 
(4). N.I.: Not Interpretable. 
Example word cluster of this definition is shown in Table~\ref{tab:lex_sem_def}.
The results are show in Table~\ref{tab:not_aligned}.
Generally,
\textsc{BertZero} covers more lexical information while
\textsc{BertLast} covers more semantics.

The results so far (Table~\ref{tab:overall}
to~\ref{tab:lex_sem_def}) confirm our hypothesis that latent
states indeed exist. 
The results also support our claim that the linguistic meanings of states do not necessarily align with well-defined annotations (even though we have selected the most aligned checkpoint over different hyperparameters and random seeds). 
Note that non-aligned states take fair portions in both
\textsc{BertZero} and \textsc{BertLast} (about 27\% and 40\% coverage respectively), 
nonetheless the annotators think most of them are still meaningful
(recall examples in Table~\ref{tab:lex_sem_def}) as about only 10+\% of
the non-aligned states are not interpretable to the annotators. 
These results highlight the difficulties faced by the mainstream supervised probing w.r.t. the use of  external supervision~\citep{chen2021probing, hall-maudslay-etal-2020-tale, hewitt-liang-2019-designing, wu-etal-2020-perturbed}
and endorse the application of unsupervised methods.

\input{fig_state_freq_sw}

\input{tab_contextualization}
\subsection{The Mechanism of Contextualization}
\label{sec:state_word:contextualization}

Now we study the mechanism of contextualization by
taking a closer look on what is encoded in \textsc{BertLast} 
but \textit{not} in \textsc{BertZero}. 
To this end, we differentiate two types of words:
(1). function words (e.g., preposition,
conjunction, determiner, punctuation .etc) whose main role 
is to help sentence construction but 
do not have concrete meanings on their own;
(2). content words (e.g., nouns, adjectives, verbs, adverbs) who
have concrete meaning.
It turns out that contextualization results in very different
behavior about the encoding of these two types.

Figure~\ref{fig:state_freq_sw} shows how
function/ content words are encoded before/ after
contextualization.
We see two effects of contextualization: 
(1) before contextualization, 
most function words are concentrated around a few head states;
after contextualization, these function words spread over the full distribution, not just head states. 
This shows that the meaning of function words is distributed from head states to all states according to their context. 
(2). before contextualization, most states are either function-only or content-only (as most bars are either orange-only or blue-only); 
after contextualization, most states contain both function and content states (as most bars have both blue and orange portions). 
This shows that the meaning of function words is entangled together with 
their neighbor content words.
Intuitively, contextualization helps function words ``receive'' meaning from their context. 

We now revisit Fig.~\ref{fig:state_word} that we 
briefly mentioned in~\cref{sec:intro}. 
Figure~\ref{fig:state_word}  is produced by
t-SNE~\citep{van2008visualizing} jointly over the states and
embeddings from \textsc{BertLast} and 
illustrates the local topology (because t-SNE preserves more local information) of the representation space.
Blue/ white circles in Fig.~\ref{fig:state_word} correspond to
blue/ orange bars in Fig.~\ref{fig:state_freq_sw} and circle size
correspond to bar height.
It directly shows how states spread over and ``receive'' meaning from their neighbor word embeddings and encode to different types of word clusters.

We further highlight certain example clusters before/ after
contextualization in Table~\ref{tab:contextualization}.
Before contextualization, we see (1). the symbol \textit{\$} does not have meaningful neighbors; (2). the suffix \textit{\#ing} and \textit{\#ed} are just ordinary subwords; (3). the word \textit{be}'s neighbor is its morphological variants.
After contextualization, we see (1). the symbol \textit{\$} encodes money;
(2). the suffix \textit{\#ing} and \textit{\#ed} encode tense; 
(3). \textit{be}'s neighbor becomes linking verbs. 
Contextualization makes these tokens ``receive'' meaning from their contexts.

%% file: fig_state_freq_sw.tex
\begin{figure*}[!t]
\small
  \centering
  \includegraphics[width=\linewidth]{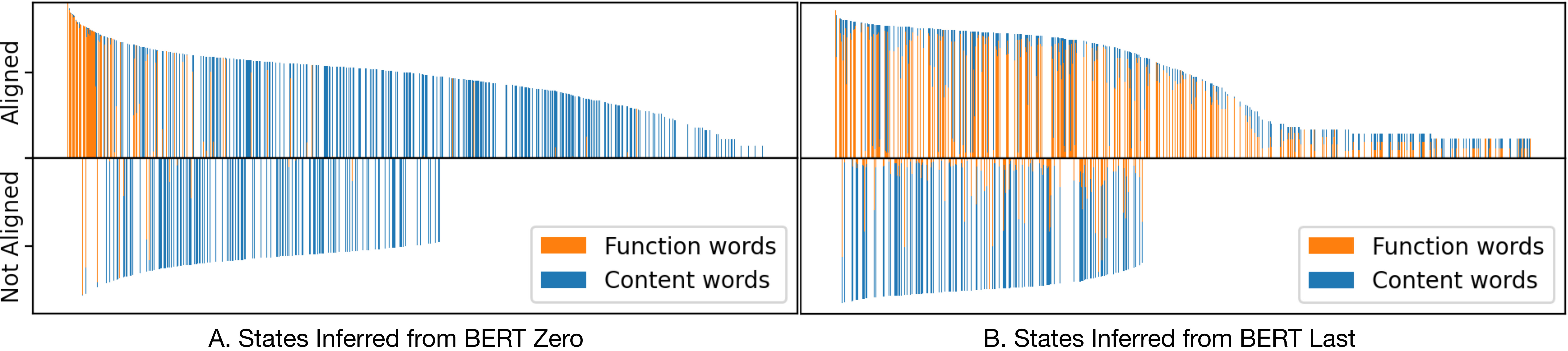}
  \caption{Mechanism of Contextualization. Each bar represents a latent state with height equals to log frequency. 
  Orange/ blue represents the portion of function/ content word corresponding to a state. 
  In \textsc{BertZero}, most function words are concentrated around head states. After contextualization (\textsc{BertLast}), function words mix with content words and spread over all states. 
  }
  \label{fig:state_freq_sw}
\end{figure*}

%% file: tab_contextualization.tex
\begin{table*}[t!]
  \small
  \caption{\label{tab:contextualization} Mechanism of contextualization. Try comparing \red{red} tokens and see their neighbor words before/ after contextualization. Tokens previously opaque (as their corresponding latent state are not meaningful) gain linguistic clarity (as their corresponding states encode linguistic constructions). 
  }
  \begin{center}
  \begin{tabular}{@{}p{1cm}p{6cm}p{6cm}@{}} 
  \toprule
  & Before Contextualization (\textsc{BertZero}) & After Contextualization (\textsc{BertLast})\\ 
  \midrule
  Symbol &  \red{\$}\sufx{2851} | size\sufx{56} | type\sufx{49} | numbers\sufx{38} | number\sufx{35} & \red{\$}\sufx{248} | money\sufx{76} | cost\sufx{64} | pay\sufx{54} | love\sufx{42} | worth\sufx{41} \\
  & \red{@}\sufx{13522} | same\sufx{2110} | ordinary\sufx{46} | average\sufx{7} & \red{@}\sufx{1184} | com\sufx{1146} | org\sufx{232} | address\sufx{91} | list\sufx{75} \\
  Prefix & re\sufx{5635} | \red{pre}\sufx{559} | mis\sufx{481} | co\sufx{258} | pr\sufx{22} &  old\sufx{657} | after\sufx{42} | recently\sufx{42} | years\sufx{36} | \red{pre}\sufx{36} \\
  & \red{un}\sufx{1922} | per\sufx{871} | di\sufx{468} | multi\sufx{237} | \#con\sufx{159} & \red{un}\sufx{1524} | in\sufx{562} | im\sufx{275} | mis\sufx{162} | con\sufx{155} | um\sufx{148} \\ 
  Suffix & \red{\#ing}\sufx{1508} | \#ting\sufx{108} | \#ley\sufx{56} | \#light\sufx{36} & \red{\#ing}\sufx{1563} | running\sufx{226} | processing\sufx{118} |  writing\sufx{98} \\
  & \red{\#ly}\sufx{1722} | dear\sufx{59} | thy\sufx{36} | \#more\sufx{15} | \#rous\sufx{9} & \red{\#ly}\sufx{983} | actually\sufx{645} | exactly\sufx{325} | simply\sufx{282}\\  
  & \#eg\sufx{404} | \red{\#ed}\sufx{385} | \#ve\sufx{189} | \#ize\sufx{183} | \#ig\sufx{164} |  & \#d\sufx{1012} | had\sufx{542} | \red{\#ed}\sufx{416} | did\sufx{320} | used\sufx{258} \\ 
  & \red{\#s}\sufx{348} | \#l\sufx{102} | \#t\sufx{98} | \#p\sufx{85} | \#m\sufx{64} | \#u\sufx{62} & \red{\#s}\sufx{1839} | files\sufx{225} | books\sufx{169} | machines\sufx{123}  \\ 
  & \red{\#s}\sufx{335} | s\sufx{333} | \#t\sufx{134} | \#p\sufx{120} | \#u\sufx{117} | it\sufx{98} & people\sufx{4481} | \red{\#s}\sufx{682} | those\sufx{361} | users\sufx{210} | folks\sufx{193} \\ 
  Lexicon & decided\sufx{250} | \red{decision}\sufx{211} | decide\sufx{189} | determine\sufx{102} | determined\sufx{99} | decisions\sufx{82} & \red{decision}\sufx{220} | position\sufx{206} | choose\sufx{155} | command\sufx{147} | actions\sufx{106} | decide\sufx{94} \\
  &\red{be}\sufx{7125} | been\sufx{1591} | being\sufx{257} | gone\sufx{1} & \red{be}\sufx{7192} | are\sufx{3099} | am\sufx{1139} |  is\sufx{165} | become\sufx{52} \\
  \bottomrule
  \end{tabular}
  \end{center}
\end{table*}

%% file: 060state_state.tex
\input{fig_transition}
\input{tab_transition}
\input{fig_bigram_freq}
Now we study the mechanism of contextualization at the phrase and sentence level. 
We first visualize the transition network 
in~\cref{sec:state_state:overall} to see the backbone of the space topology.
Then we show how sentences are constructed as traversals over the discovered latent state network (\cref{sec:state_state:sentence_traversal}). 

\subsection{Overall State Transition Topology}
\label{sec:state_state:overall}
We visualize the induced state-state network 
in Fig.~\ref{fig:transition} using t-SNE again (this time without word embeddings).
Blue circles represent states with more content words while 
white circles represent states with more function words.
Circle size represents state frequency. 
To see how states transit to each other, 
we compute the state transition statistics from the state sequences decoded from the validation dataset.
The transition histogram is also shown in Fig.~\ref{fig:bigram_dist}.
We use blue edges to denote frequent (stronger) transitions and yellow edges to denote less frequent (weaker) transitions. 
Table~\ref{tab:transition} shows example transitions and their corresponding word bigram occurrences.
In Fig.~\ref{fig:transition}A we see: 
(1). both nodes and edges follow a long-tail distribution:
 there are few frequent nodes/ edges taking the head portion of the distribution, and many infrequent nodes/
edges taking the tail portion of the distribution. 
Note that the yellow background in Fig.~\ref{fig:transition}A consists of many weak edges. 
(2). frequent states are more inter-connected and tail states are more spread.
Fig.~\ref{fig:transition}B zooms in the top states, and we see function states usually as the \textit{hub} of the edges. 
This is also evidenced in Table~\ref{tab:transition}, as we can see
many different content words transit to the function word \textit{to} (e.g., free-to, willing-to, go-to, went-to), and 
the function word ``to'' can transit to other content words (e.g., to-buy, to-sell, to-build).
Here the state encoding \textit{to} is a hub connecting other states and words.

Figure~\ref{fig:bigram_dist} shows transition distribution.
The bars here correspond to edges in Fig.~\ref{fig:transition}.
Color denote the portion of function/ content words 
(node color in Fig.~\ref{fig:transition}) and 
height correspond to edge color in Fig.~\ref{fig:transition}.
We observe:
(1). transitions are usually mixtures of function and content states. 
(2). contextualization makes function words less concentrated around head transitions (in Fig.~\ref{fig:bigram_dist} left, \textsc{BertLast} has less orange portion than \textsc{BertZero}), and more spread within the distribution (in Fig.~\ref{fig:bigram_dist} right, \textsc{BertLast} has a longer tail of orange bars than \textsc{BertZero}).

\subsection{Sentence Encoding as Traversals over the Latent Network}
\label{sec:state_state:sentence_traversal}
Now we can finally put everything together and reach the most
important discovery of this paper: 
the latent mechanism of sentence
representation within the topology.
This mechanism consists of four steps, and is illustrated in Fig.\ref{fig:traversal}.
Step 1: there exist function states that correspond to specific function words (as is evidenced in Fig.~\ref{fig:state_freq_sw}).
Step 2: there exist content states that correspond to content words with similar lexical/ syntactic/ semantic meaning (as is evidenced in Table~\ref{tab:overall} to~\ref{tab:lex_sem_def}, Fig.~\ref{fig:state_word}). 
Step 3: transitions between function and content states correspond to meaningful phrase constructions (as is evidenced in Fig.~\ref{fig:transition} and~\ref{fig:bigram_dist}, Table~\ref{tab:transition}). 
Step 4: a traversal of states encodes a sentence within the space (this is a corollary combining step 1-3). 
Fig.~\ref{fig:traversal} shows sentences sharing overlapped traversal.
State transition chains encode the sentence templates and  state-word emissions fill in the content. 
When the transition chains of two sentences overlap, 
the two sentences tend to be syntactically similar.

\input{fig_traversal}

%% file: fig_transition.tex
\begin{figure*}[!t]
\small
  \centering
  \includegraphics[width=\linewidth]{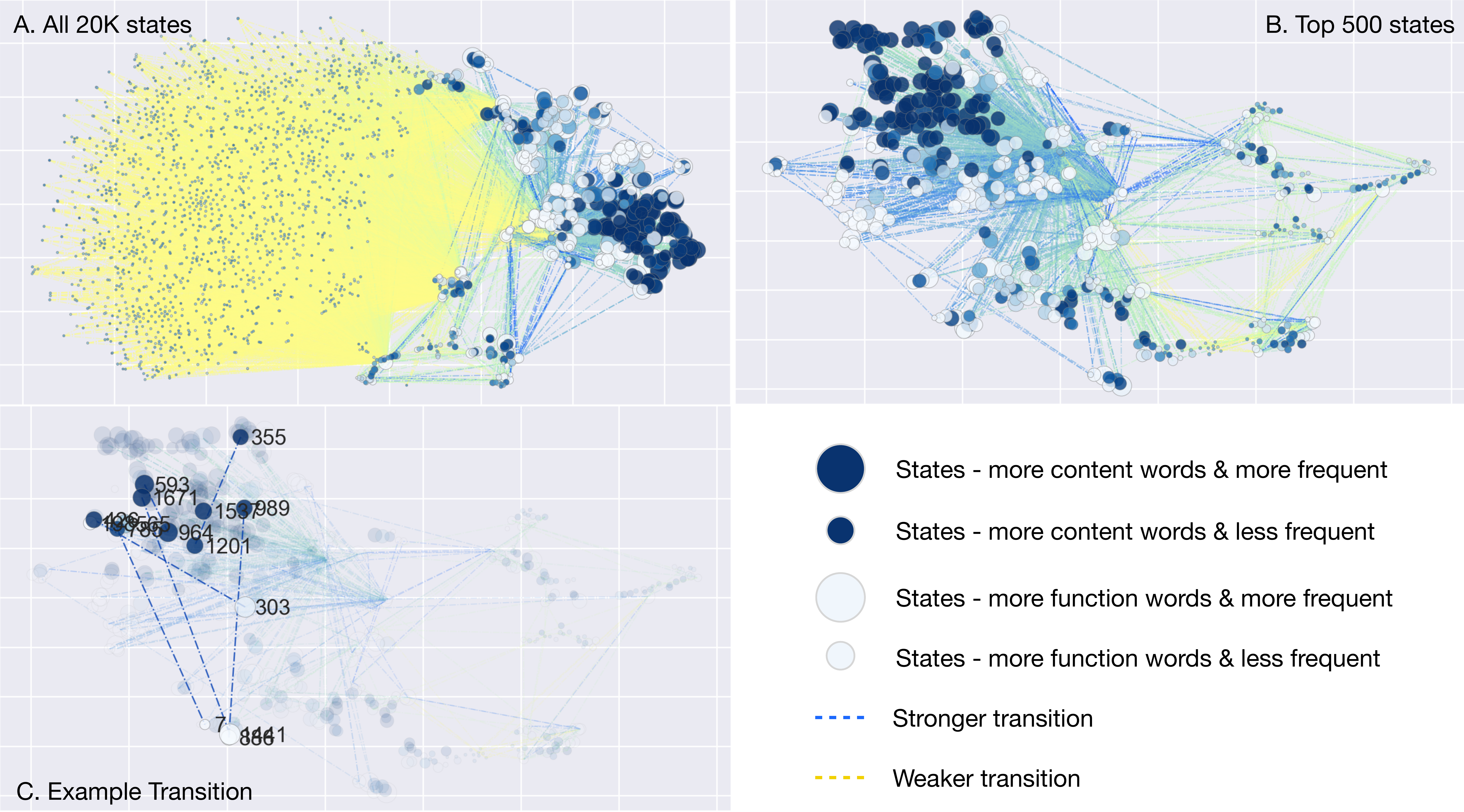}
  \caption{
  A: State transitions as the backbone of the representation space. 
  Tail states are more spread and transitions to tail states mostly are from head states (yellow edges). 
  Head states are more inter-connected (blue edges). 
  Blue edges denote more frequent (stronger) transitions and yellow edges denote less frequent (weaker) transitions.
  B: transitions between top states, function states usually (white circles) serve as hubs of transitions (many white circles at the center). 
  C. Example transitions, see Table~\ref{tab:transition} for their interpretations. 
  }
  \label{fig:transition}
\end{figure*}

%% file: tab_transition.tex
\begin{table*}[t!]
  \caption{\label{tab:transition}  Example state transitions. Function words (like ``to'') serve as hubs that attract content words of similar meanings to be within the same latent state. Subscript numbers denote occurrence. 
  }
  \begin{center}
  \small
  \begin{tabular}{@{}p{1.4cm}ll@{}} 
  \toprule
   Transition & Example states from \textsc{BertLast} & Explanation \\ 
   \midrule
  \multicolumn{3}{@{}l}{\footnotesize{Function word + content word}} \\
    886-989 & to-buy\sufx{21} | to-sell\sufx{13} | to-build\sufx{5} | to-purchase\sufx{5} | to-create\sufx{4} | to-produce\sufx{3}  & to do sth. \\ 
    1671-1441 & free-to\sufx{14} | willing-to\sufx{9} | hard-to\sufx{6} | easy-to\sufx{4} | happy-to\sufx{3} | safe-to\sufx{2} & adjective + to \\ 
    785-7 & go-to\sufx{19} | going-to\sufx{10} | went-to\sufx{5} | trip-to\sufx{4} | moved-to\sufx{3} | come-to\sufx{2} & movement + to \\
    785-565 & come-out\sufx{9} | come-up\sufx{6} | went-up\sufx{4} | go-down\sufx{3} | went-out\sufx{3} | went-back\sufx{1} & move + direction
    \\
    426-198 & caused-by\sufx{9} | \#ed-by\sufx{8} | \#ted-by\sufx{5} | produced-by\sufx{4} | made-by\sufx{4} | driven-by\sufx{3} & passive voice \\ 
    303-426 & is-made\sufx{2} | be-converted\sufx{2} | been-formed\sufx{2} | are-formed\sufx{2} | been-developed\sufx{1} & passive voice \\ 
    \multicolumn{3}{@{}l}{\footnotesize{Content word + content word}} \\
    1537-1537 & ms-windows\sufx{13} | source-code\sufx{5} | windows-program\sufx{3} | operating-system\sufx{3} & computers \\ 
    355-1201 & three-years\sufx{7} | five-years\sufx{3} | 24-hours\sufx{3} | ten-years\sufx{2} | 21-days\sufx{2} & time \\
    593-964 & image-processing\sufx{2} | meter-reading\sufx{1} | missile-spotting\sufx{1} | speed-scanning\sufx{1} & v.ing as noun 
    \\
  \bottomrule
  \end{tabular}
  \end{center}
\end{table*}

%% file: fig_bigram_freq.tex
\begin{figure*}[!t]
\small
  \centering
  \includegraphics[width=\linewidth]{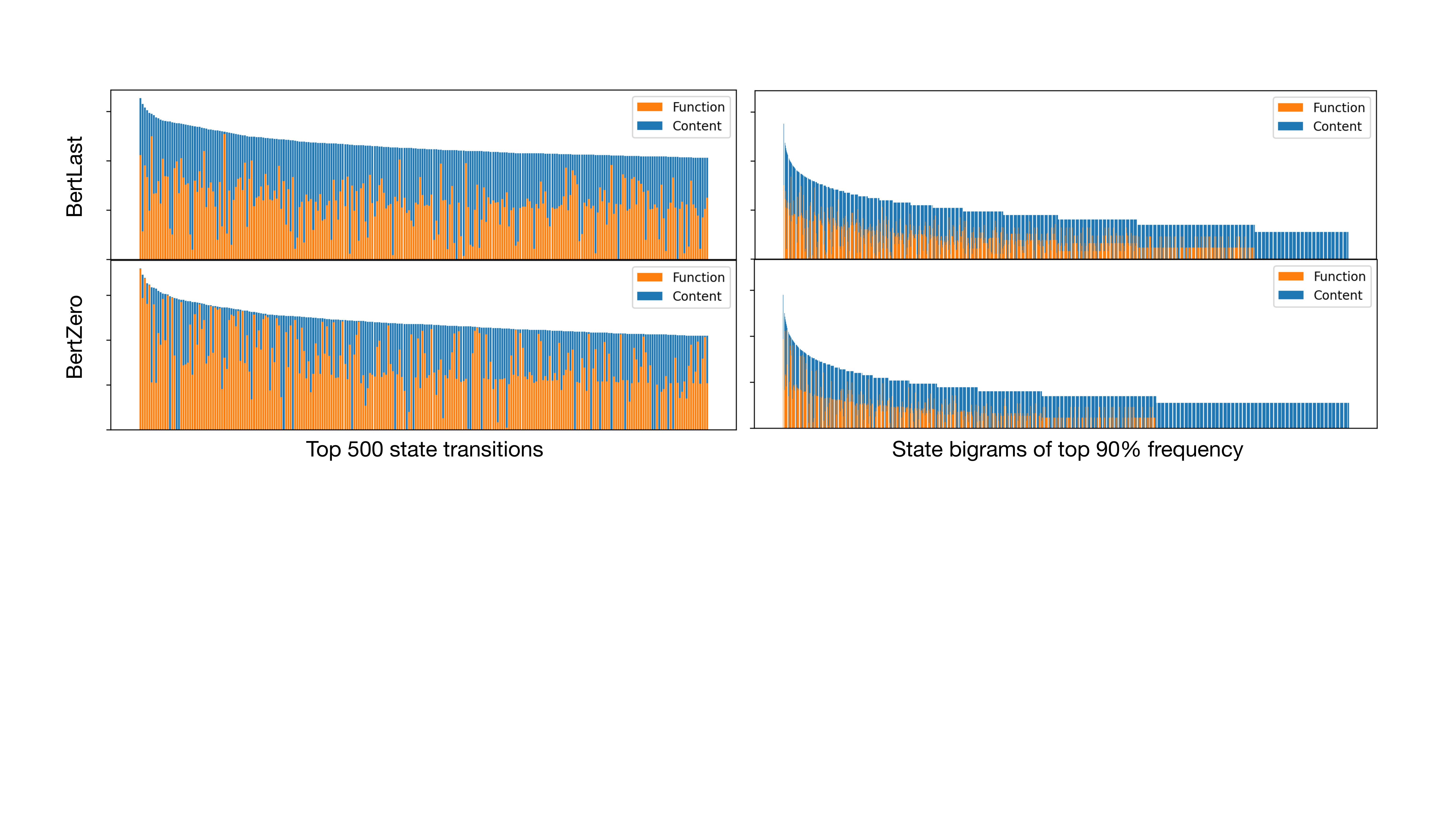}
  \caption{State transition distribution. Again, function words exist more at top transitions in \textsc{BertZero}. After contextualization (\textsc{BertLast}), they become more spread within the distribution.
  }
  \label{fig:bigram_dist}
\end{figure*}

%% file: fig_traversal.tex
\begin{figure*}[!t]
\small
  \centering
  \includegraphics[width=\linewidth]{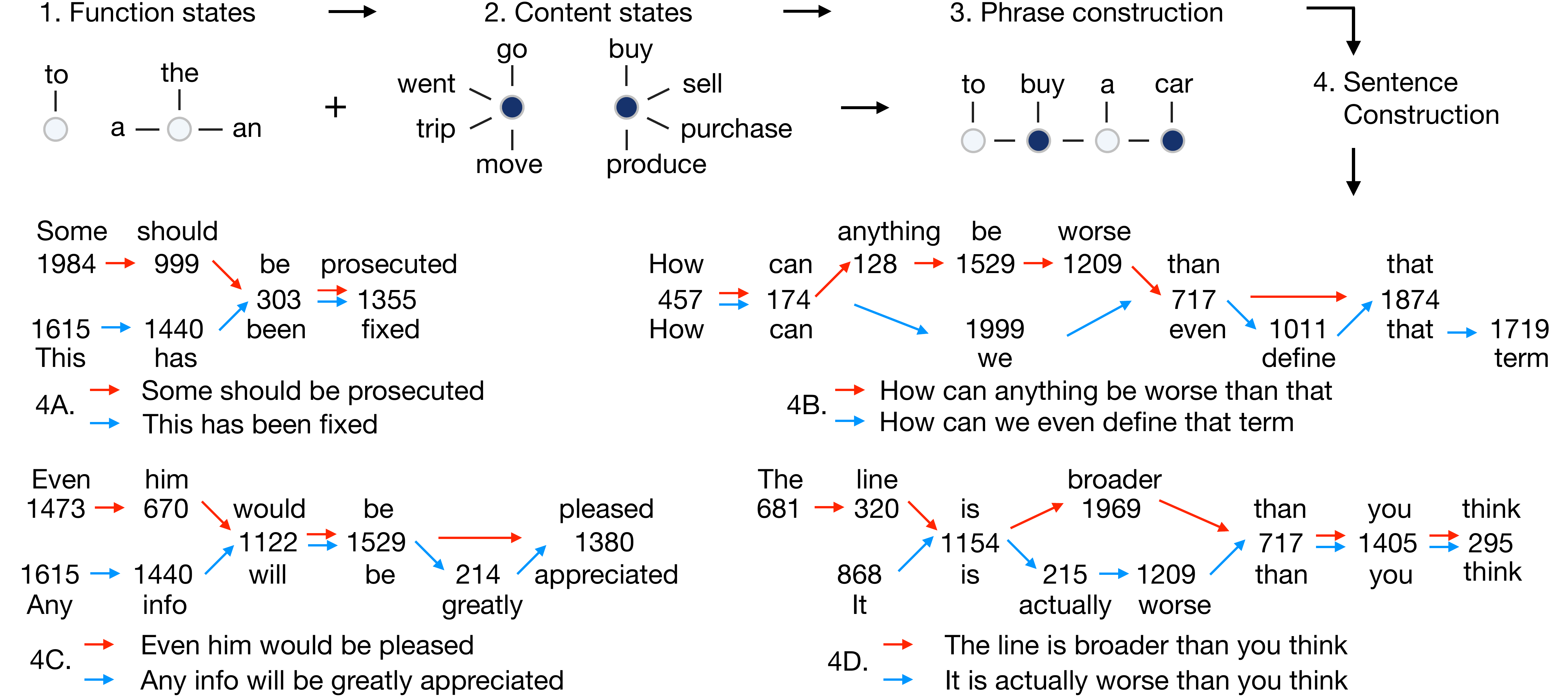}
  \caption{
  Four steps illustration of the mechanism about how sentences are represented as a traversal 
  over the latent state network. 
  Numbers mean latent state index. 
  Structurally similar sentences share overlapped paths of latent states. 
  }
  \label{fig:traversal}
\end{figure*}

%% file: 070conclusion.tex
In this work, we discover a latent state network intrinsically within
the representation space of contextualized representations.
Our analytics starts from the hypothesis that there exists a latent network of states that summarize the representation space topology.
We verify such states indeed exist, and they do not necessarily align to existing well-defined annotations (\cref{sec:state_word:decompozation}). 
Then we reveal the mechanism of contextualization by showing how words within states ``receive'' meaning from their context and become interpretable after contextualization (\cref{sec:state_word:contextualization}). 
We further study how state transitions mark the backbone of the representation space and encode meaningful phrase constructions (\cref{sec:state_state:overall}).
Finally, combining the state-word and state-state topology, 
we reach the latent mechanism about how sentences are encoded as a traversal over the state network (\cref{sec:state_state:sentence_traversal}). 

Due to the space and time limit, our results have the following major limitation: 
(1). our analysis is more about the \textit{topological} structure of the space (i.e., how nodes are connected), but less about the \textit{distance} structure (i.e., how far one node is from another), while the later is also an important geometric property. 
(2). techniques used in our analysis is more about \textit{local} topology (e.g., neighbor words around a state), but less about the \textit{global} topology (e.g., how states and words form hierarchies). 
(3). there are evidences~\citep{cai2021isotropy} that the autoregressive-styled transformer (GPT2 or the decoder of T5) have different topologies than bidirectional-styled transformer (BERT and encoder of T5), and we only explore BERT. 
We leave the exploration of these directions to future work.

Finally, we note that although the literature on Bertology is rich, our understanding of the model behavior is still far from complete, especially for properties discovered with unsupervised methods. 
We hope this work deepens the understanding of language models, encourages unsupervised analytics, and inspires new modeling techniques based on the topology of the representations.

%% file: app_main.tex
\textbf{Appendix Table of Content}
\begin{itemize}
\item Section~\ref{app:exp_details}. More experimental details.
\item  Section~\ref{app:state_word_details}. More word cluster examples.
\item Section~\ref{app:state_state_details}. More state transition examples.
\end{itemize}

\input{app_tab_funcwords}
\section{Experimental Details}
\label{app:exp_details}

\subsection{Training Details}

\textbf{$\beta$ Parameter}\quad To get meaningful convergence of the latent space without posterior collapse, 
the most sensitive parameter is the $\beta$ parameter in Eq.5 for entropy regularization.
We need to set $\beta$ in the correct range. 
A large $\beta$ force the posterior to collapse to a uniform prior, while a 
small $\beta$ encourages the posterior to collapse to a Dirac distribution.
To search for a meaningfull $\beta$, we start with different order of magnitudes: 0.1, 0.01, 0.001, 0.0001, 0.00001. 
We find out 0.001 and 0.01 gives the best performance. 
Then we search within this range using linear interpolation: 0.001, 0.0025, 0.005, 0.0075, 0.01, and find out 0.005 gives the best performance. 
So we set $\beta$ to be 0.005. 

\textbf{Robustness to Random seeds}\quad 
The results reported in Section 5 are searched over three random seeds and we choose the run that has the largest number of aligned tags to the union of the five types of exising annotation (i.e., POS, ENT, DEP, CCG, BCN). 
The differences between random seeds are not large.
For example in Table 1 (main paper) the number of not aligned states with \textsc{BertLast} is 1069, and other runs produce about 1059-1069. 
Generally, our results are robust enough to random seeds. 
We further note that the example word clusters, the mechanism of contextualization, and state transitions are also consistent w.r.t. random seeds. 
We can observe similar word clusters and state transitions in Fig.1 and 3 and Table 4 and 5 (main paper) with different random seeds. 
This is to say, our discovered topology is \textit{not} specific to the chosen random seeds, it is an intrinsic property of the BERT representation space.

\textbf{Hardware}\quad Generally, one Nvidia 2080Ti will 11G memory would suffice   one single run.
In our hyperpameter searching, we usually run multiple instances (say, with 8 2080Tis) at the same time.
For groups trying to reproduce our results, we would say two 2080Tis would suffice, yet the more the better. 

\subsection{Function Words Used in This Work}
Table~\ref{tab:app:funcwords} shows the functions words used in this work. 
This is a list of stopwords defined by NLTK\footnote{\url{https://www.nltk.org/book/ch02.html}}, and we reuse them here. 
Other words not in this list are viewed as content words.

\subsection{Visualization Procedure}
\label{sec:app:visualization}
This section describes how we produce the tSNE visualization in Figure 1 and 3 (main paper). 
We use the sklearn implementation~\footnote{\url{https://scikit-learn.org/stable/modules/generated/sklearn.manifold.TSNE.html}}
 of tSNE.
For Fig.1, the word embeddings are obtained by 
sampling 8,000 contextualized embeddings from the full word occurrences in 
the 20News dev set. 
Then we put the sampled word embeddings and the 2,000 states into the tSNE 
function. 
The perplexity is set to be 30. 
An important operation we use, for better separating the states from each other,
it to manually set the distances between states to be large, otherwise 
the states would be concentrated in a sub-region, rather than spread over words.
Fig.3 is produced similarly, except we do not use 
word embeddings as background, and change the perplexity to be 5.
All our decisions of hyperparameters are for the purpose of clear visualization which includes reducing overlapping, overcrowding, and other issues. 
We further note that tSNE itself is a visualization for local topology.
Here no single visualization method can reveal the full 
global structure of high-dimensional data, and any projection to 2-D plain inevitably
induces information loss.
We leave the investigation of better visualization methods to future work.

\input{app_tab_state_word_1}
\input{app_tab_state_word_2}
\section{State-Word Topology, More Examples}
\label{app:state_word_details}
See Table~\ref{tab:app:state_word_more_1} and~\ref{tab:app:state_word_more_2} for more state-word examples. 

\input{app_tab_state_state_1}
\input{app_tab_state_state_2}
\section{State-State Topology, More Examples}
\label{app:state_state_details}
See Table~\ref{tab:app:state_state_more_1} and~\ref{tab:app:state_state_more_2} for more state transition examples.

%% file: app_tab_funcwords.tex
\begin{table*}[t]
  \begin{center}
    \begin{tabular}{@{}lllllllll@{}}   
    \toprule
    i & me & my & myself & we & our & ours & ourselves & you\\
your & yours & yourself & yourselves & he & him & his & himself & she\\
her & hers & herself & it & its & itself & they & them & their\\
theirs & themselves & what & which & who & whom & this & that & these\\
those & am & is & are & was & were & be & been & being\\
have & has & had & having & do & does & did & doing & a\\
an & the & and & but & if & or & because & as & until\\
while & of & at & by & for & with & about & against & between\\
into & through & during & before & after & above & below & to & from\\
up & down & in & out & on & off & over & under & again\\
further & then & once & here & there & when & where & why & how\\
all & any & both & each & few & more & most & other & some\\
such & no & nor & not & only & own & same & so & than\\
too & very & s & t & can & will & just & don & should\\
now & &&&&&&&\\ 
    \bottomrule
    \end{tabular}
  \end{center}
  \caption{\label{tab:app:funcwords} Function words used in the experiments.  
  }
  \end{table*}

%% file: app_tab_state_word_1.tex
\begin{table*}[t]
  \small
  \begin{center}
    \begin{tabular}{@{}lp{13cm}@{}}   
    \toprule
    State id & Word - Occurrence \\ 
    \midrule 
    1724 & \#\#s 5092  ;  \#\#es 239  ;  \#\#os 153  ;  \#\#is 140  ;  \#\#as 76  ;  \#\#z 70  ;  \#\#rs 54  ;  \#\#gs 48  ;  states 46  ;  \#\#t 46  ;  \#\#p 46  ;  \#\#ps 43  ;  \#\#ms 40  ;  \#\#ns 35  ;  \#\#e 34  ;  \#\#r 31  ;  \#\#i 29  ;  \#\#ss 27 \\ \cmidrule{2-2}
    476 & \#\#ed 609  ;  \#\#d 437  ;  \#\#ted 156  ;  based 144  ;  caused 95  ;  \#\#ized 75  ;  made 61  ;  lost 61  ;  built 60  ;  \#\#able 58  ;  supported 57  ;  expressed 56  ;  occupied 54  ;  defined 54  ;  created 54 \\ \cmidrule{2-2}
    254 & david 521  ;  john 404  ;  mike 249  ;  steve 245  ;  dave 234  ;  michael 223  ;  robert 214  ;  jim 204  ;  mark 194  ;  bob 173  ;  paul 143  ;  james 142  ;  bill 133  ;  tom 127  ;  andrew 122  ;  peter 121 \\ \cmidrule{2-2}
    710 & problem 599  ;  killed 278  ;  death 243  ;  kill 235  ;  problems 211  ;  error 187  ;  crime 181  ;  murder 176  ;  sin 175  ;  genocide 172  ;  evil 140  ;  bad 129  ;  hate 112  ;  pain 110  ;  massacre 97 \\ \cmidrule{2-2}
    1972 &  \#\#s 649  ;  turks 222  ;  armenians 206  ;  jews 186  ;  keys 171  ;  muslims 151  ;  arabs 123  ;   \#\#ms 122  ;  christians 93  ;   \#\#es 73  ;  countries 71  ;  villages 63  ;  children 61  ;  colors 61  ;  guns 61 \\ \cmidrule{2-2}
    403 & god 1917  ;   \#\#a 751  ;  world 730  ;  ca 347  ;  country 292  ;  usa 289  ;   \#\#o 284  ;  group 280  ;   \#\#u 277  ;  uk 244  ;  groups 133  ;   \#\#ga 126  ;   \#\#g 125  ;  europe 124  ;  heaven 120  ;  hell 119 \\ \cmidrule{2-2}
    1494 & problem 257  ;  problems 87  ;  agree 67  ;  discussion 41  ;  issue 39  ;  argument 34  ;  deal 34  ;  issues 31  ;  disagree 20  ;  conflict 16  ;  arguments 16  ;  case 15  ;  relationship 14  ;  agreement 14 \\ \cmidrule{2-2}
    1419 & real 512  ;  general 231  ;  major 203  ;  specific 157  ;  main 156  ;  actual 133  ;  important 116  ;  full 109  ;  total 106  ;  serious 99  ;  absolute 88  ;  basic 87  ;  great 84  ;  personal 83  ;  true 76 \\ \cmidrule{2-2}
    305 &  \#\#s 1336  ;  opinions 340  ;  problems 154  ;  rules 140  ;  views 113  ;  things 104  ;  laws 99  ;  events 97  ;  actions 94  ;  issues 94  ;  values 93  ;  arguments 90  ;  cases 82  ;  rates 78  ;  comments 78 \\ \cmidrule{2-2}
    555 & subject 2349  ;  key 596  ;  name 551  ;  number 472  ;  address 300  ;  size 242  ;  numbers 167  ;  value 149  ;  color 141  ;  important 134  ;  speed 122  ;  names 110  ;  level 106  ;  count 84 \\ \cmidrule{2-2}
    1556 &  \#\#ing 1282 ; running 188 ;  \#\#ting 149 ;  \#\#ng 104 ;  \#\#ling 92 ; processing 87 ; killing 83 ;  \#\#ding 71 ; calling 70 ;  \#\#ring 69 ; getting 66 ; reading 61 ;  \#\#izing 61 ;  \#\#ping 60  \\ \cmidrule{2-2}
    472 & re 3084 ; com 2685 ; e 1034 ;  \#\#e 892 ; internet 271 ;  \#\#o 185 ;  \#\#te 140 ; de 117 ;  \#\#re 101 ;  \#\#com 98 ;  \#\#ne 71 ;  \#\#r 71 ;  \#\#er 70 ; org 70 ;  \#\#me 70 ; net 68 ; se 62 \\ \cmidrule{2-2}
    1488 & human 344 ; turkish 255 ; political 246 ; moral 217 ; religious 186 ; legal 161 ; jewish 159 ; israeli 143 ; sexual 128 ; social 121 ; physical 119 ; personal 114 ; civil 104 ; international 103  \\ \cmidrule{2-2}
    1410 & simple 292 ; standard 213 ; normal 210 ; reasonable 208 ; fine 192 ; nice 191 ; interesting 138 ; good 128 ; correct 127 ; perfect 121 ; stupid 117 ;  \#\#able 106 ; proper 105 ; true 91 \\ \cmidrule{2-2}
    243 &  \#\#ly 929 ; probably 311 ; clearly 254 ; completely 231 ; obviously 229 ; certainly 222 ; directly 186 ; easily 179 ; apparently 176 ; generally 162 ; simply 154 ; necessarily 153 ; unfortunately 143 \\ \cmidrule{2-2}
    1418 & never 919 ; little 48 ; cannot 46 ; none 37 ; nobody 35 ; ever 31 ; nothing 27 ; without 25 ; neither 19 ; didn 14 ; ve 11 ; hardly 9 ; always 8 ; zero 8 ; m 7 ; non 6 ; doesn 6 \\ \cmidrule{2-2}
    1246 &  \#\#t 1242 ;  \#\#p 853 ; net 568 ; phone 395 ; call 381 ; bit 348 ;  \#\#net 323 ;  \#\#l 225 ; bat 135 ; line 123 ;  \#\#g 111 ;  \#\#d 101 ; network 94 ;  \#\#it 93 ; tel 87 ; telephone 83 ;  \#\#m 79 \\ \cmidrule{2-2}
    935 & long 682 ; hard 600 ; big 570 ; bad 518 ; good 483 ; great 404 ; fast 216 ; large 185 ;  \#\#y 160 ; short 158 ; quick 154 ; little 150 ; strong 145 ; high 114 ; nice 113 ; hot 107 \\ \cmidrule{2-2}
    659 & cs 474 ; jp 468 ; cc 315 ; ms 291 ; cd 217 ; cm 194 ; dc 181 ;  \#\#eg 158 ; bt 142 ; ps 125 ; ds 123 ; mc 120 ; pc 110 ; nc 110 ; gt 107 ; gs 107 ; rs 103 ; ss 103 ; bc 83 ; cb 81 \\ \cmidrule{2-2} 
    1214 & space 182 ; nasa 170 ; orbit 169 ; motif 136 ; moon 103 ; planet 95 ; prism 94 ; lunar 92 ; venus 86 ; saturn 85 ; spacecraft 85 ; earth 80 ; shuttle 76 ; satellite 73 ; mars 70 \\ \cmidrule{2-2} 
    654 & around 351 ; within 234 ; behind 137 ; across 102 ; regarding 81 ; among 73 ; throughout 70 ; beyond 57 ; near 53 ; along 53 ; inside 47 ; outside 44 ; past 40 ; concerning 38 ; following 37 \\  \cmidrule{2-2} 
    127 &  \#\#er 473 ; everyone 390 ;  \#\#r 319 ; user 270 ; anyone 259 ; host 180 ; friend 161 ; one 150 ; server 136 ; everybody 109 ; fan 104 ; manager 103 ; nobody 101 ; player 93 ; guy 84 \\ \cmidrule{2-2} 
    145 & believe 798 ; hope 421 ; evidence 379 ; claim 308 ; test 219 ; assume 201 ; proof 148 ; argument 145 ; prove 133 ; check 127 ; claims 106 ; suspect 104 ; doubt 95 ; guess 93 ; assuming 85 \\ 
    \bottomrule
    \end{tabular}
  \end{center}
  \caption{\label{tab:app:state_word_more_1} More state-word examples 
  }
  \end{table*}

%% file: app_tab_state_word_2.tex
\begin{table*}[t]
  \small
  \begin{center}
    \begin{tabular}{@{}lp{13cm}@{}}   
    \toprule
    State id & Word - Occurrence \\ 
    \midrule 
    1572 & president 378 ; clinton 293 ; \#\#resh 268 ; myers 165 ; attorney 84 ; general 79 ; morris 78 ; smith 76 ; paul 75 ; bush 74 ; manager 64 ; hitler 56 ; \#\#ey 52 ; \#\#i 48 ; \#\#man 45 \\ \cmidrule{2-2}
    964 & cut 132 ; plug 121 ; break 73 ; thread 73 ; cable 63 ; hole 59 ; holes 54 ; chip 49 ; fix 48 ; clutch 48 ; stick 46 ; connector 42 ; blow 42 ; box 41 ; screw 40 ; pin 40 ; hit 40 \\ \cmidrule{2-2}
    1756 & see 1721 ; look 858 ; seen 618 ; read 302 ; saw 274 ; display 205 ; image 199 ; looks 197 ; looking 196 ; looked 188 ; screen 177 ; watch 161 ; view 153 ; monitor 149 ; images 132 \\ \cmidrule{2-2}
    585 & day 779 ; sun 686 ; \#\#n 556 ; today 310 ; night 276 ; week 269 ; days 264 ; city 161 ; morning 145 ; sunday 125 ; \#\#en 117 ; year 105 ; \#\#net 96 ; n 92 ; \#\#on 89 ; weeks 87 ; month 73  \\ \cmidrule{2-2}
    66 & power 433 ; control 399 ; state 347 ; virginia 202 ; mode 182 ; process 162 ; effect 139 ; period 118 ; action 117 ; authority 91 ; function 87 ; position 84 ; \#\#v 82 ; force 78 \\ \cmidrule{2-2}
    1240 & first 1443 ; always 602 ; ever 559 ; never 361 ; ago 321 ; often 319 ; sometimes 284 ; usually 203 ; early 195 ; last 192 ; every 175 ; later 171 ; soon 155 ; recently 146 ; past 145 \\ \cmidrule{2-2}
    865 & faith 383 ; religion 377 ; believe 211 ; atheist 205 ; \#\#ism 164 ; islam 159 ; \#\#sm 158 ; religious 145 ; morality 137 ; belief 126 ; font 115 ; language 114 ; truth 92 ; logic 90 \\ \cmidrule{2-2}
    467 & \#\#u 1203 ; point 784 ; happen 234 ; place 201 ; happened 183 ; colorado 141 ; happens 139 ; points 132 ; wait 126 ; ground 94 ; site 94 ; center 86 ; position 78 ; situation 78 ; 1993 76 \\  \cmidrule{2-2}
    203 & ca 273 ; pub 177 ; \#\#u 143 ; dod 143 ; au 141 ; mit 138 ; ma 132 ; \#\#si 129 ; sera 121 ; des 113 ; fi 75 ; isa 70 ; il 58 ; ny 58 ; po 56 ; la 53 ; tar 48 ; lee 47 ; ti 47 \\ \cmidrule{2-2}
    371 & drug 212 ; drugs 177 ; food 145 ; health 130 ; medical 121 ; disease 117 ; diet 115 ; cancer 113 ; aids 98 ; homosexuality 96 ; sex 83 ; homosexual 82 ; medicine 82 ; hiv 78 ; treatment 77 \\ \cmidrule{2-2}
    1683 & money 479 ; cost 307 ; pay 274 ; issue 212 ; problem 186 ; matter 175 ; worth 175 ; care 153 ; costs 146 ; tax 108 ; expensive 102 ; responsible 96 ; risk 96 ; spend 95 ; insurance 94 \\ \cmidrule{2-2}
    1856 & whole 379 ; entire 196 ; full 58 ; every 49 ; everything 42 ; together 37 ; everyone 25 ; \#\#u 17 ; rest 14 ; \#\#up 13 ; \#\#ed 13 ; away 10 ; always 10 ; top 10 ; open 10 ; \#\#s 9 \\ \cmidrule{2-2}
    1584 & university 1064 ; government 886 ; law 769 ; science 482 ; \#\#u 412 ; research 312 ; history 290 ; laws 165 ; study 125 ; policy 105 ; court 103 ; scientific 98 ; physics 97 ; constitution 93 \\ \cmidrule{2-2}
    1514 & life 663 ; live 363 ; security 192 ; exist 188 ; peace 180 ; dead 175 ; living 164 ; existence 157 ; body 157 ; lives 153 ; exists 137 ; privacy 128 ; death 126 ; die 121 ; safety 112 \\ \cmidrule{2-2}
    1208 & atheist 165 ; font 144 ; bio 119 ; bb 111 ; homosexual 109 ; \#\#group 99 ; sy 94 ; mormon 75 ; fed 72 ; manual 56 ; posting 52 ; spec 52 ; \#\#s 50 ; \#\#eri 49 ; auto 42 ; pointer 41 ; handgun 37 \\ \cmidrule{2-2}
    837 & another 974 ; last 774 ; next 581 ; else 578 ; second 498 ; others 472 ; third 124 ; first 82 ; final 69 ; later 60 ; rest 49 ; future 47 ; 2nd 44 ; latter 41 ; previous 40 ; elsewhere 33 \\ \cmidrule{2-2}
    1291 & lines 1848 ; read 701 ; writes 502 ; line 376 ; book 319 ; books 244 ; write 203 ; written 177 ; text 171 ; reading 157 ; wrote 144 ; article 107 ; quote 92 ; writing 86 ; paper 84 \\ \cmidrule{2-2}
    1656 & agree 182 ; solution 165 ; advice 128 ; opinion 110 ; interface 104 ; response 88 ; suggestions 80 ; recommend 75 ; alternative 75 ; discussion 71 ; offer 71 ; argument 70 ; application 69 \\ \cmidrule{2-2}
    1874 & apple 405 ; chip 386 ; disk 373 ; fbi 289 ; encryption 197 ; \#\#eg 171 ; hardware 166 ; nsa 154 ; ram 154 ; algorithm 134 ; tape 129 ; nasa 119 ; chips 111 ; ibm 100 ; floppy 98 \\ \cmidrule{2-2}
    1966 & stanford 269 ; washington 177 ; russian 156 ; cleveland 141 ; berkeley 137 ; california 131 ; chicago 105 ; \#\#co 96 ; turkey 95 ; york 83 ; boston 74 ; bosnia 73 ; soviet 71 ; russia 71 \\ \cmidrule{2-2}
    603 & file 682 ; list 526 ; article 501 ; card 424 ; bill 237 ; board 196 ; book 191 ; box 180 ; package 140 ; page 139 ; directory 119 ; section 118 ; group 114 ; library 90 ; files 83 \\ \cmidrule{2-2}
    1401 & done 644 ; didn 41 ; perform 35 ; performed 32 ; accomplish 25 ; accomplished 16 ; could 14 ; \#\#d 11 ; conduct 10 ; happen 10 ; say 10 ; committed 9 ; finish 9 ; completed 9 ; conducted 8 \\ \cmidrule{2-2}
    460 & clip 186 ; \#\#op 175 ; com 162 ; news 162 ; posts 109 ; works 106 ; micro 68 ; sim 66 ; share 66 ; \#\#yp 58 ; net 58 ; wire 54 ; \#\#os 48 ; power 43 ; es 40 ; flop 39 ; mac 39 ; tool 39 \\ 
    \bottomrule
    \end{tabular}
  \end{center}
  \caption{\label{tab:app:state_word_more_2} More state-word examples, continued.
  }
  \end{table*}

%% file: app_tab_state_state_1.tex
\begin{table*}[t]
  \small
  \begin{center}
    \begin{tabular}{@{}lp{13cm}@{}}   
    \toprule
    Transition & Bigram - Occurrence \\ 
    \midrule 
    1843-990 & is-that 68 ; fact-that 51 ; so-that 50 ; think-that 46 ; note-that 41 ; say-that 39 ; sure-that 38 ; believe-that 35 ; out-that 33 ; know-that 32 ; seems-that 28 ; mean-that 26 \\ \cmidrule{2-2}
    1010-1016 & instead-of 40 ; amount-of 33 ; lot-of 26 ; form-of 23 ; lack-of 19 ; institute-of 16 ; case-of 15 ; capable-of 14 ; amounts-of 13 ; out-of 12 ; years-of 12 ; department-of 11 ; terms-of 11 \\ \cmidrule{2-2}
    960-458 & up-to 56 ; down-to 29 ; access-to 25 ; according-to 24 ; due-to 22 ; go-to 17 ; response-to 14 ; subject-to 13 ; related-to 13 ; reference-to 13 ; as-to 12 ; lead-to 12 ; reply-to 12  \\ \cmidrule{2-2}
    441-698 & have-to 139 ; going-to 116 ; seem-to 114 ; seems-to 68 ; supposed-to 40 ; need-to 30 ; had-to 26 ; used-to 22 ; want-to 20 ; seemed-to 15 ; tend-to 14 ; appears-to 13 ; likely-to 13 ; appear-to 12 \\ \cmidrule{2-2}
    1712-698 & trying-to 67 ; try-to 46 ; able-to 43 ; like-to 41 ; hard-to 32 ; seem-to 22 ; seems-to 22 ; want-to 21 ; tend-to 21 ; willing-to 18 ; tried-to 16 ; enough-to 14 ; attempt-to 13 ; continue-to 12 \\ \cmidrule{2-2} 
    1814-1666 & about-it 91 ; of-it 71 ; with-it 42 ; to-it 29 ; for-it 27 ; do-it 26 ; on-it 24 ; have-it 12 ; understand-it 11 ; doing-it 9 ; know-it 9 ; see-it 8 ; call-it 8 ; believe-it 8 ; \#\#ing-it 7 ; fix-it 6 \\ \cmidrule{2-2} 
    1295-523 & problem-with 32 ; deal-with 32 ; do-with 28 ; up-with 16 ; problems-with 15 ; came-with 13 ; comes-with 13 ; along-with 12 ; work-with 12 ; contact-with 11 ; wrong-with 10 ; agree-with 10 ; disagree-with 9 \\ \cmidrule{2-2}  
    628-150 &  based-on 65 ; depending-on 23 ; is-on 13 ; \#\#s-on 11 ; down-on 9 ; effect-on 9 ; are-on 8 ; working-on 8 ; effects-on 7 ; activities-on 7 ; depend-on 7 ; be-on 6 ; run-on 6 ; depends-on 6 \\ \cmidrule{2-2}  
    477-1414 & have-to 117 ; going-to 45 ; is-to 37 ; had-to 32 ; decided-to 12 ; need-to 11 ; has-to 11 ; having-to 9 ; required-to 9 ; willing-to 8 ; how-to 8 ; ,-to 7 ; reason-to 7 ; forced-to 7  \\ \cmidrule{2-2}  
    477-1277 & is-to 74 ; have-to 43 ; had-to 20 ; used-to 17 ; required-to 14 ; going-to 14 ; ,-to 13 ; need-to 13 ; as-to 12 ; order-to 11 ; needed-to 11 ; \#\#s-to 10 ; be-to 10 ; decided-to 10 \\ \cmidrule{2-2}  
    145-461 & believe-that 70 ; claim-that 24 ; evidence-that 18 ; assume-that 17 ; hope-that 15 ; belief-that 11 ; sure-that 9 ; prove-that 9 ; assuming-that 8 ; argue-that 8 ; likely-that 7 ; claims-that 7 \\ \cmidrule{2-2}  
    278-217 & know-of 22 ; end-of 16 ; out-of 14 ; think-of 13 ; \#\#s-of 10 ; accuracy-of 8 ; top-of 7 ; friend-of 6 ; copy-of 6 ; heard-of 6 ; one-of 4 ; middle-of 4 ; version-of 4 ; beginning-of 4 ; aware-of 4 \\ \cmidrule{2-2}  
    1820-276 & come-out 30 ; came-out 17 ; coming-out 14 ; put-out 12 ; get-out 11 ; find-out 10 ; check-out 9 ; turns-out 7 ; found-out 7 ; turn-out 7 ; turned-out 7 ; comes-out 7 ; go-out 6 ; \#\#ed-out 6 \\ \cmidrule{2-2}  
    1142-461 & is-that 17 ; fact-that 15 ; understand-that 12 ; see-that 11 ; realize-that 11 ; noted-that 8 ; says-that 8 ; note-that 7 ; read-that 7 ; forget-that 6 ; out-that 6 ; shows-that 6 \\ \cmidrule{2-2}  
    1010-1998 & lot-of 34 ; set-of 26 ; bunch-of 24 ; lots-of 22 ; series-of 13 ; number-of 10 ; thousands-of 10 ; hundreds-of 10 ; plenty-of 10 ; full-of 7 ; pack-of 7 ; list-of 6 ; think-of 5 \\ \cmidrule{2-2}  
    1125-843 & of-a 124 ; is-a 86 ; for-a 84 ; to-a 50 ; s-a 16 ; be-a 14 ; ,-a 11 ; as-a 7 ; was-a 5 ; on-a 5 ; with-a 4 ; am-a 3 ; about-a 3 ; in-a 2 ; into-a 2 ; were-a 2 ; its-a 1 ; surrounding-a 1 \\ \cmidrule{2-2}  
    476-1654 & written-by 13 ; \#\#d-by 11 ; caused-by 8 ; \#\#ed-by 8 ; produced-by 6 ; followed-by 6 ; defined-by 4 ; committed-by 4 ; hit-by 4 ; supported-by 4 ; led-by 4 ; explained-by 4 ; run-by 4 \\ \cmidrule{2-2}  
    1812-837 & the-other 86 ; the-next 77 ; the-last 62 ; the-second 48 ; the-first 14 ; the-latter 10 ; the-third 9 ; the-latest 7 ; the-rest 6 ; the-previous 6 ; the-final 5 ; the-fourth 3 ; the-nearest 3 \\ \cmidrule{2-2}  
    1938-145 & i-believe 128 ; i-hope 66 ; i-suspect 28 ; i-assume 24 ; i-doubt 18 ; i-suppose 11 ; i-guess 11 ; i-expect 8 ; i-think 7 ; i-imagine 6 ; i-feel 5 ; i-trust 4 ; i-gather 3 ; i-bet 2 \\ \cmidrule{2-2}  
    1820-1856 & pick-up 14 ; come-up 12 ; came-up 11 ; stand-up 11 ; set-up 11 ; bring-up 8 ; show-up 8 ; comes-up 7 ; screwed-up 7 ; give-up 6 ; wake-up 6 ; speak-up 5 ; look-up 5 ; back-up 5  \\ \cmidrule{2-2}  
    1417-979 & more-than 163 ; better-than 33 ; less-than 13 ; faster-than 12 ; greater-than 11 ; longer-than 8 ; \#\#er-than 7 ; larger-than 6 ; worse-than 6 ; higher-than 6 ; slower-than 6 ; easier-than 4 \\ \cmidrule{2-2}  
    111-111 & of-the 75 ; to-the 34 ; for-the 23 ; on-the 14 ; with-the 12 ; about-the 7 ; part-of 7 ; in-the 5 ; into-the 5 ; like-the 4 ; out-of 4 ; at-the 4 ; by-the 3 ; '-s 3 ; as-the 2 \\ \cmidrule{2-2}  
    1579-654 & talking-about 45 ; talk-about 25 ; concerned-about 14 ; worried-about 9 ; know-about 8 ; stories-about 7 ; worry-about 7 ; talked-about 6 ; rumours-about 5 ; news-about 5 ; feel-about 5 ; care-about 4 \\
    \bottomrule
    \end{tabular}
  \end{center}
  \caption{\label{tab:app:state_state_more_1} State transition examples, with function words
  }
  \end{table*}

%% file: app_tab_state_state_2.tex
\begin{table*}[t]
  \small
  \begin{center}
    \begin{tabular}{@{}lp{13cm}@{}}   
    \toprule
    Transition & Bigram - Occurrence \\ 
    \midrule 
    371-371 & health-care 14 ; side-effects 8 ; im-\#\#mun 4 ; infectious-diseases 4 ; yeast-infections 4 ; \#\#thic-medicine 3 ; treat-cancer 3 ; health-insurance 3 ; barbecue-\#\#d 3 ; hiv-infection 3 ; yeast-syndrome 3 \\ \cmidrule{2-2}
    1214-1214 & orbit-\#\#er 14 ; astro-\#\#physics 7 ; lunar-orbit 7 ; space-shuttle 7 ; earth-orbit 5 ; pioneer-venus 5 ; space-station 5 ; space-\#\#lab 4 ; lunar-colony 4 ; orbit-around 3 ; space-tug 3 ; space-\#\#flight 3  \\ \cmidrule{2-2}
    716-1556 & mail-\#\#ing 15 ; fra-\#\#ering 12 ; \#\#mina-\#\#tion 9 ; bash-\#\#ing 7 ; \#\#dal-\#\#izing 6 ; \#\#ras-\#\#ing 5 ; \#\#band-\#\#ing 4 ; \#\#ress-\#\#ing 4 ; cab-\#\#ling 4 ; adapt-\#\#er 4 ; cluster-\#\#ing 4 ; sha-\#\#ding 4 \\ \cmidrule{2-2}
    931-931 & gamma-ray 17 ; lead-acid 9 ; wild-corn 4 ; mile-long 3 ; smoke-\#\#less 3 ; drip-\#\#py 2 ; diamond-stealth 2 ; cold-fusion 2 ; 3d-wire 2 ; acid-batteries 2 ; schneider-stealth 2 ; quantum-black 2 \\ \cmidrule{2-2}
    1488-1488 & law-enforcement 17 ; national-security 5 ; cold-blooded 4 ; health-care 4 ; human-rights 4 ; im-\#\#moral 4 ; prophet-\#\#ic 4 ; social-science 3 ; ethnic-\#\#al 3 ; turkish-historical 3\\ \cmidrule{2-2}
    1246-1246 & bit-\#\#net 35 ; tel-\#\#net 12 ; use-\#\#net 7 ; phone-number 7 ; dial-\#\#og 6 ; \#\#p-site 5 ; phone-calls 5 ; bit-block 5 ; net-\#\#com 4 ; bat-\#\#f 4 ; \#\#t-\#\#net 4 ; phone-call 4 ; arc-\#\#net 3\\ \cmidrule{2-2}
    1556-1556 & abu-\#\#sing 5 ; \#\#dal-\#\#izing 4 ; obey-\#\#ing 4 ; robb-\#\#ing 3 ; \#\#ov-\#\#ing 3 ; dial-\#\#ing 3 ; contend-\#\#ing 3 ; \#\#upt-\#\#ing 3 ; rough-\#\#ing 3 ; contact-\#\#ing 3 ; bash-\#\#ing 3 ; favor-\#\#ing 2 \\ \cmidrule{2-2}
    202-202 &  western-reserve 21 ; case-western 20 ; ohio-state 19 ; united-states 10 ; penn-state 5 ; african-american 5 ; north-american 5 ; middle-eastern 5 ; polytechnic-state 4 ; north-carolina 4\\ \cmidrule{2-2}
    1912-1912 & world-series 9 ; home-plate 7 ; division-winner 4 ; runs-scored 4 ; batting-average 4 ; game-winner 3 ; sports-\#\#channel 3 ; plate-umpire 3 ; baseball-players 3 ; league-baseball 3 \\ \cmidrule{2-2}
    1461-1461 &  \#\#l-bus 5 ; bit-color 5 ; 3d-graphics 4 ; \#\#p-posting 3 ; computer-graphics 3 ; wire-\#\#frame 3 ; bit-graphics 2 ; \#\#eg-file 2 ; access-encryption 2 ; \#\#frame-graphics 2 ; file-format 2 \\ \cmidrule{2-2}
    123-123 & health-care 10 ; high-school 6 ; es-\#\#crow 6 ; key-es 5 ; high-power 4 ; local-bus 4 ; low-level 4 ; high-speed 3 ; minor-league 2 ; health-service 2 ; regular-season 2 ; mother-\#\#board 2\\ \cmidrule{2-2}
    1702-1702 & mile-\#\#age 8 ; engine-compartment 5 ; semi-auto 5 ; manual-transmission 5 ; drive-power 4 ; door-car 3 ; passenger-cars 3 ; sports-car 3 ; shaft-drive 3 ; mini-\#\#van 3 ; speed-manual 3\\ \cmidrule{2-2}
    1874-1874 & floppy-disk 11 ; jp-\#\#eg 11 ; encryption-algorithm 8 ; \#\#per-chip 7 ; \#\#mb-ram 7 ; \#\#ga-card 6 ; encryption-devices 5 ; silicon-graphics 4 ; disk-drive 4 ; floppy-drive 4\\ \cmidrule{2-2}
    1208-1064 & atheist-\#\#s 43 ; homosexual-\#\#s 12 ; fed-\#\#s 9 ; libertarian-\#\#s 8 ; \#\#eri-\#\#s 7 ; \#\#tile-\#\#s 7 ; azerbaijani-\#\#s 6 ; \#\#tar-\#\#s 6 ; mormon-\#\#s 5 ; sniper-\#\#s 5 ; physicist-\#\#s 4\\ \cmidrule{2-2}
    1710-1710 & power-supply 5 ; atomic-energy 4 ; water-ice 4 ; power-cord 4 ; \#\#com-telecom 3 ; light-pollution 3 ; light-bulb 3 ; radio-station 3 ; radio-\#\#us 3 ; air-conditioning 3 ; light-\#\#wave 2 \\ \cmidrule{2-2}
    1080-1080 & public-access 19 ; via-anonymous 5 ; private-sector 5 ; available-via 4 ; general-public 4 ; community-outreach 4 ; public-domain 3 ; personal-freedom 3 ; private-property 3 ; private-activities 3\\ \cmidrule{2-2}
    254-1572 & jimmy-carter 9 ; george-bush 9 ; bill-clinton 8 ; bryan-murray 4 ; joe-carter 4 ; henry-spencer 4 ; bill-james 4 ; janet-reno 4 ; craig-holland 4 ; clayton-cramer 4 ; \#\#zie-smith 4\\ \cmidrule{2-2}
    1571-1571 & ms-windows 24 ; windows-nt 12 ; ibm-pc 10 ; ms-\#\#dos 7 ; unix-machine 6 ; microsoft-windows 5 ; windows-applications 4 ; run-windows 3 ; apple-monitor 3 ; mac-\#\#s 3 ; desktop-machine 3 \\ \cmidrule{2-2}
    66-66 & \#\#ian-1919 3 ; energy-signature 2 ; charlotte-\#\#sville 2 ; environment-variables 2 ; duty-cycle 2 ; second-period 2 ; spin-state 2 ; power-consumption 2 ; inter-\#\#mission 2 ; power-play 2\\ \cmidrule{2-2}
    1683-1683 &  worth-\#\#while 4 ; nominal-fee 4 ; get-paid 3 ; risk-factors 3 ; scholarship-fund 2 ; cost-\$ 2 ; tax-dollars 2 ; beneficial-item 2 ; bank-account 2 ; take-responsibility 2 \\ \cmidrule{2-2}
    1579-1579 &  m-sorry 5 ; news-reports 4 ; heard-anything 4 ; ran-\#\#ting 3 ; short-story 3 ; news-reporters 3 ; press-conference 3 ; heard-something 3 ; tv-coverage 2 ; horror-stories 2 ; heard-horror 2\\ \cmidrule{2-2}
    1656-1656 & urbana-champaign 3 ; peace-talks 3 ; acceptable-solutions 2 ; marriage-partner 2 ; intercontinental-meetings 2 ; interested-parties 2 ; conference-calls 2 ; handle-conference 2 ; cooperative-behaviour 2  \\ 
    \bottomrule
    \end{tabular}
  \end{center}
  \caption{\label{tab:app:state_state_more_2} State transition examples, without function words
  }
  \end{table*}